\def\year{2015}
\documentclass[letterpaper]{article}
\usepackage{aaai}
\usepackage{times}
\usepackage{courier}
\usepackage{paralist}
\usepackage[pdftex]{graphicx}
\usepackage{algorithm}
\usepackage{algpseudocode}
\usepackage{amsmath,algorithm}
\usepackage{amsmath,amssymb}
\usepackage{color}
\usepackage{url}

\usepackage{amsmath}
\usepackage{amsmath,amssymb}
\usepackage{mathtools}
\usepackage{commath}
\usepackage{amsmath}
\newenvironment{Proof}[1][Proof.]{\begin{trivlist}
\item[\hskip \labelsep {\bfseries #1}]}{\end{trivlist}}
\usepackage{amsfonts}
\usepackage{makecell}
\usepackage{listings}
\usepackage{rotating}
\usepackage{tabularx}
\usepackage{booktabs}
\usepackage{rotating}
\usepackage{tabularx}
\usepackage{multicol}
\usepackage{multirow}
\begin{document}
%
%
\title
{
Learning and Tuning Meta-heuristics in Plan Space Planning
}

\author{ Shashank Shekhar and Deepak Khemani\\
Department of Computer Science \& Engineering\\
IIT Madras, Chennai\\
Tamil Nadu, India 600 036\\
\url{{sshekhar,khemani}@cse.iitm.ac.in}
}
\maketitle
\begin{abstract}

In recent years, the planning community has observed that techniques for learning heuristic functions 
have yielded improvements in performance. 
One approach is to use offline learning to learn predictive models from existing heuristics in a domain dependent manner. 
These learned models are deployed as new heuristic functions.
%
%
The learned models can in turn be tuned online using a domain independent error correction 
approach to further enhance their informativeness. 
The online tuning  approach is domain independent but instance specific, 
and contributes to improved performance for individual instances as planning proceeds. 
Consequently it is more effective in larger problems.

In this paper, we mention two approaches applicable in Partial Order Causal Link (POCL) Planning that is also known as 
Plan Space Planning. 
First, we endeavour to enhance the performance of a POCL planner by giving an algorithm for supervised learning.
~Second, we then discuss an online error minimization approach in POCL framework to minimize the 
step-error associated with the offline learned models thus enhancing their informativeness.
Our evaluation shows that the learning approaches scale up the performance of the planner over standard benchmarks, specially for larger problems.  
%
%
\end{abstract}
\section{Introduction}
In the recent International Planning Competitions (IPC) state-space based and total-order planners
like LAMA~\cite{richter2010lama}, 
Fast Downward Stone Soup~\cite{helmert2011fast},
and 
Fast Downward Stone Soup 2014~\cite{rogerfast14} have performed well. 
These planners are very efficient, generate consistent states fast, and use powerful state-based heuristics.
But they often commit too early to ordering of actions, giving up on flexibility in the generated plans.
\textcolor{black}
{
In contrast, the POCL framework~\cite{ColesCFL10} generates more flexible plans, but in general 
is computationally more intensive than the state-space based approaches.
}
%
%
%
The POCL framework has found applicability in multi-agent planning~\cite{Kvarnstrom11}
and temporal planning~\cite{BentonCC12}.

%
Researchers have recently investigated the applicability of state-space heuristic learning 
approaches~\cite{sapena2014combining,ColesCFL10} in POCL planning.
This revival of interest is due to the idea of delayed commitment of
RePOP~\cite{nguyen2001reviving} and VHPOP~\cite{younes2003vhpop}.
%
%
%
In this paper  we further investigate the adaptation of state space approaches in POCL planners yielding quality plans over the same or even larger problems.
%
%
%
In general, due to the diverse nature of planning problems characterized by the degree of interactions between subgoals, 
a heuristic function does not always work well in all planning domains.
Various techniques have been devised to increase the informativeness of heuristics in the state space arena.
One approach strengthens the delete relaxation heuristic by incrementing lower bounds to 
tighten the admissibility of the heuristic, repeatedly by solving relaxed versions of a planning problem~\cite{Haslum12}.
%
%
%
%
%
%
%
%
In another approach, 
to circumvent the trade-offs of combining multiple heuristics, 
a decision rule is used for selecting 
a heuristic function in a given state. 
An active online learning technique is applied to learn a model for that given decision rule~\cite{DomshlakKM10}. 

%
We present a couple of machine learning techniques,  
influenced  from~\cite{arfaee2011learning,thayer2011learning,samadi2008learning,us2013learning}, 
which improve the effectiveness of heuristics in the POCL framework. 
First, we apply domain wise regression techniques in a supervised manner, using existing POCL heuristics as features. 
For target generation we use our planner named RegPOCL.
This is based on VHPOP and uses grounded actions to speed up the planning process.
Next, we further give another technique for POCL framework 
which enhances the informativeness of these offline learned models. 
This technique is an adapted version of an online heuristic adjustment 
approach based on temporal difference (TD) learning~\cite{sutton1988learning,thayer2011learning}.
We extend this domain independent approach for learning instance specific details, 
which corrects the error associated with the learned models, thus making them more informed.
\textcolor{black}
{
The RegPOCL planner employs these two approaches and evaluation shows that it is more efficient on the benchmarks.
} 
We have confined the evaluation to non-temporal STRIPS domains.

The rest of the paper is structured as follows. 
After looking at the motivation for exploring the POCL approach, 
we describe the learning approaches used along with the experiment design. 
These are followed by the experimental results and the concluding discussion.
\section{POCL Planning}
A POCL planner starts search with a null partial plan and progresses over a space of partial plans,
by adding a resolver to a partial plan to remove a flaw.
We use heuristics from~\cite{younes2003vhpop} for selecting the most adverse flaw to resolve in the selected 
partial plan.
%
%
%
The POCL framework has certain advantages over forward state-space search {(FSSS)}.  
{FSSS} has the problem of premature commitment to an ordering between two actions which reduces the plan flexibility.  
It does so to avoid any mutual interference between actions, though there may not be any. 
POCL planning avoids committing unnecessarily to ordering actions.  
%
%
%


%
\textcolor{black}
{
FSSS faces problems while solving instances with deadlines.
Deadlines may arise within the interval(s) of one or more durative actions. 
In general, the actions may produce some delayed effects, and this may have ramifications on deadlines as well, 
which creates deadlines relative to their starting points~\cite{ColesCFL10}. 
FSSS also suffers from significant backtracking as it may explore all possible plan permutations 
in the absence of effective guidance.
}
The POCL framework also has several advantages in temporal planning,
specially in planning with complex temporal constraints beyond actions with duration. 
%
%
These limitations of {FSSS} motivate us to explore the POCL framework.    
\subsubsection{Example.}
Suppose we are required to add four actions [$a_1$, $a_2$, $a_3$, $a_4$] to a plan, 
where $a_2$ is dependent on $a_1$ and $a_4$ is dependent on $a_3$. 
There is no interference between any two actions apart from the above dependencies. 
In this case, FSSS gives an ordering or timestamp [0, 1, 2, 3], with a makespan 4,
whilst the delayed commitment strategy would give more choices with flexibility in the orderings like 
[2, 0, 1, 3] and [0, 2, 1, 3]. If parallel execution is allowed, makespan would be 2.
If another action $a_5$, which is dependent on $a_3$, has to be introduced in the plan then FSSS will allot it a timestamp 4, whereas delayed-commitment strategy could allot it 1. 

However, if we ignore the absence of the flexibility and action parallelism in {FSSS}, 
it is very fast in recovering from a situation that would arise due to committing to some wrong choices during planning. 
{FSSS} has the advantage of faster search state generation and powerful state-based heuristics.  
\section{Learning Approaches Used}
We propose a two fold approach to learn better heuristic functions. 
First, existing heuristic function are combined by a process of offline learning that generates learned predictive models. 
This is followed by an online technique of adjusting the step-error associated with these models during partial plan refinement.
We divide this section into two parts: the first describes the offline learning techniques to perform regression, and 
the second the technique of further strengthening the informativeness of a learned predictive model.  
\subsection{Offline Learning}
Offline learning is an attractive approach because generating training sets in most planning domains is 
a fast and simple process. 
%
%
%
%
%
The learning process has three phases: 
\begin{inparaenum}[(\itshape i\upshape)]
\item dataset preparation, 
\item training, and 
\item testing.
\end{inparaenum}
The training instances gathered by solving small problems become the inputs to the used regression techniques
(\emph{e.g.} linear regression and M5P, described later), 
which learn effective ways of combining existing heuristic functions. 
The testing phase is used to validate the best ways of combining the known heuristic functions.  
%
%
Algorithm~1, described below, embodies the complete training process. 
\begin{algorithm}
\caption{The algorithm used during training phase}
\begin{algorithmic}[1]
\State  \textbf{Input} 
\State \hspace{0.15in} \emph{AS} - Attribute Selection; $T$ - Training Dataset; 
\State \hspace{0.15in} $S$ - Problem Set; $L$ - Learning Technique;
\State \hspace{0.15in} $H$ - Heuristic Set; \emph{RegPOCL} - The Planner.
\State \textbf{Output}
\State \hspace{0.15in} $\mathrm{M}:\mathrm{T}\rightarrow\mathbb{R}$~~~// \emph{A learned predictive model}
\Procedure {Training-Algorithm}{$\emph{AS}$, $T$, $L$}
\State $T$ $\leftarrow$ $\phi$~~~// \emph{Domain specific}.
\State $T$ $\leftarrow$ \Call{Dataset-prep}{\emph{RegPOCL}, $S$, $H$, $T$}
\State Training Instances $\leftarrow$ Apply($\emph{AS}$, $T$)
\State \Return $M$ $\leftarrow$ Apply($L$, Training Instances)
\EndProcedure
\Statex
\Procedure{Dataset-prep}{\textit{{RegPOCL}}, $S$, $H$, $T$} 
\State \hspace{0.0in} $F$ - Check; $\mathcal{T}$ - Target Value;
\State \hspace{0.0in} $\Pi$ - A set of seed partial plans.
	\For{\textbf{each} $p$ $\in$ $S$}
		\State $\Pi$ $\leftarrow$ Null partial plan of $p$
    	\For{a random $sp$ $\in$ $\Pi$} 
    		\State ($F$, $\mathcal{T}$, $\Pi_{loc}$) $\leftarrow$ {Solve(\emph{RegPOCL}, $sp$, $H$)}
    		\If{$F$}~~~// $sp$ \emph{refines completely}. 
				\State $\Pi$ $\leftarrow$ $\Pi$ $\cup$ $\Pi_{loc}$
				\State Ins $\leftarrow$ {Comp-inst}({\emph{RegPOCL}, $H$, $sp$, $\mathcal{T}$)}
				\State $T$ $\leftarrow$ $T$ $\cup$ Ins
    		\EndIf
    	\EndFor~~~// \emph{Bounded number of iterations}.
    	%
	\EndFor
    \State \Return $T$
\EndProcedure
\end{algorithmic}
\end{algorithm}
\subsubsection{Dataset Preparation}
%
%
%
The procedure \sc{dataset-prep()} \normalfont
Line 13 in Algorithm~1 is used to solve a set ($S$) of planning problems. 
We consider only small problems that are gathered from each planning domain selected from previous IPCs.
The output of Algorithm~1 is a trained predictive model (as shown in line~6).
We consider each problem from $S$ for the dataset preparation in each domain (line~16).
In this algorithm, a seed partial plan is a new partial plan that gets generated due to
a possible refinement of the selected partial plan. 
We select a seed $sp$ from a set of seed partial plans $\Pi$ (line~18). 
$sp$ will be provided to {RegPOCL} for its further refinements.
If {RegPOCL} is able to generate one consistent solution by refining $sp$ completely using \emph{Solve()} function (line~19), 
then the flag $F$ will be true. 
We capture the newly generated partial plans in local instance of $\Pi$ called $\Pi_{loc}$. 
The target $\mathcal{T}$ captures the number of new actions that get added in $sp$.
$\mathcal{T}$ is calculated when the planner refines $sp$ completely using heuristics from $H$.
Note that, $\mathcal{T}$ is not a heuristic value but the number of new actions added during the refinement process. 
The value of $\mathcal{T}$ is also the plan length found which might not be optimal.
Since $sp$ is refined completely, $\Pi_{loc}$ is updated to $\Pi$ (line~21). 
Line~22 computes a training instance $\mathrm{Ins}$, using \emph{Comp-inst()} function. 
For a given $sp$, the planner generates a training instance of the form 
$t(sp)$ = $\langle\langle h_1(sp)$, $h_2(sp),$ $ h_3(sp),$ $h_4(sp),$ $h_5(sp),$ $h_6(sp)\rangle,$ $\mathcal{T}\rangle$,
where $h_1$ to $h_6$ are the feature heuristics.
%
%
%
%
%
To maintain consistency, we update the training set $T$ (line~23) only when {RegPOCL} refines the current seed $sp$ completely.
If complete refinement was not possible, all new seeds from $\Pi_{loc}$ are dropped, 
even though it might be the case that $\Pi_{loc}$ contains some consistent seeds particularly in the case of time out. 
%
%
To maintain diversity in $T$, for a given domain we randomly select a fixed number of seeds for 
the complete refinement process (line~18). 

We execute Algorithm~1 once for each selected domain with a given set of feature heuristics. 
Note that learning does not guarantee optimal predictive models even though optimal targets 
have been used in the training~\cite{us2013learning}. 
Algorithm~1 hunts for a well informed heuristic using learning and does not bother about its admissibility.
Since the state-of-the-art POCL heuristics are not optimal in nature~\cite{younes2003vhpop},
the usage of {RegPOCL} for generating training instances should not affect the performance of {RegPOCL} on large 
problems in the testing phase.
The selection of {RegPOCL} for generating training sets might deteriorate the actual target values, 
as the targets calculated by {RegPOCL} are not optimal in general. 
Thus there is a possibility of learning inaccurate predictive models in the training phase,
which might reduce the informativeness of the models.
We enhance the informativeness of the models by correcting the step-error associated with them using an
online heuristic tuning approach. 
%
%
%
%
\subsubsection{Training}
Once Algorithm~1 generates enough number of training instances for a given domain, it moves to the next step (line~10).
%
%
We define a regression model $\mathrm{M}:\mathrm{T}\rightarrow\mathbb{R}$, 
where $\mathrm{T}$ is a training set and $\mathbb{R}$ is a set of real numbers.
%
%
Following the general model training strategy, 
we use WEKA~\cite{Hall2009} to remove irrelevant or redundant attributes (line~10) from the training set.
This reduces the effort of the planner because the planner must calculate the 
selected feature heuristics at each step of the planning process. 
Next, we apply model training process (line~11).
We feed the training instances to different machine learning approaches to learn different predictive models. 
\subsubsection{Testing}
We test the predictive models on large problems.
The models are directly compared to the current best heuristics in the POCL framework. 
For using machine learning approaches in planning efficiently, we select the best learned regression models
and test the efficiency of {RegPOCL} by using them over different benchmarks. 
These models help in selecting the most suitable partial plan for refinement.
Offline learning learns a better model in terms of search guidance and accuracy than
online learning~\cite{samadi2008learning,thayer2011learning}.
An offline learned predictor is more accurate than an online one because 
in the offline case a complete training set is available.
%
%
Another alternative to the above approaches would be bootstrapping methods~\cite{arfaee2011learning},
\textcolor{black}
{
where a set of problems is solved using a base heuristic within a specified time limit. 
Later, the solutions obtained for learning are used to generate a new more informed heuristic function.
}
%
%
%
%
%
%
%
%
%
\subsection{Online Error Adjustment of A Predictive Model}
The offline approach assumes that the learned knowledge can effectively be transferred from 
the training instances to the test instances.
This is not always true as the planning problems are not always similar even though they belong to the same domain. 
Also, the training instances are required before the training process.
For a planning domain, during the dataset preparation, calculation of features is not computationally hard,
but calculating actual number of new actions $\mathcal{(T)}$ needed for $sp$ is relatively expensive. 
%
%
%
Online learning attempts to avoid these limitations. 
%
%
In our hybrid approach we use small instances for offline training, thus saving on time. 
This is followed up with online learning to improve the heuristic estimate on-the-fly.
The online error tuning is based on temporal difference (TD) learning~\cite{sutton1988learning}. 
%
%
%
%

\begin{figure}[ht]
\centering
     \includegraphics[width=0.45 \textwidth, decodearray={3 5}]{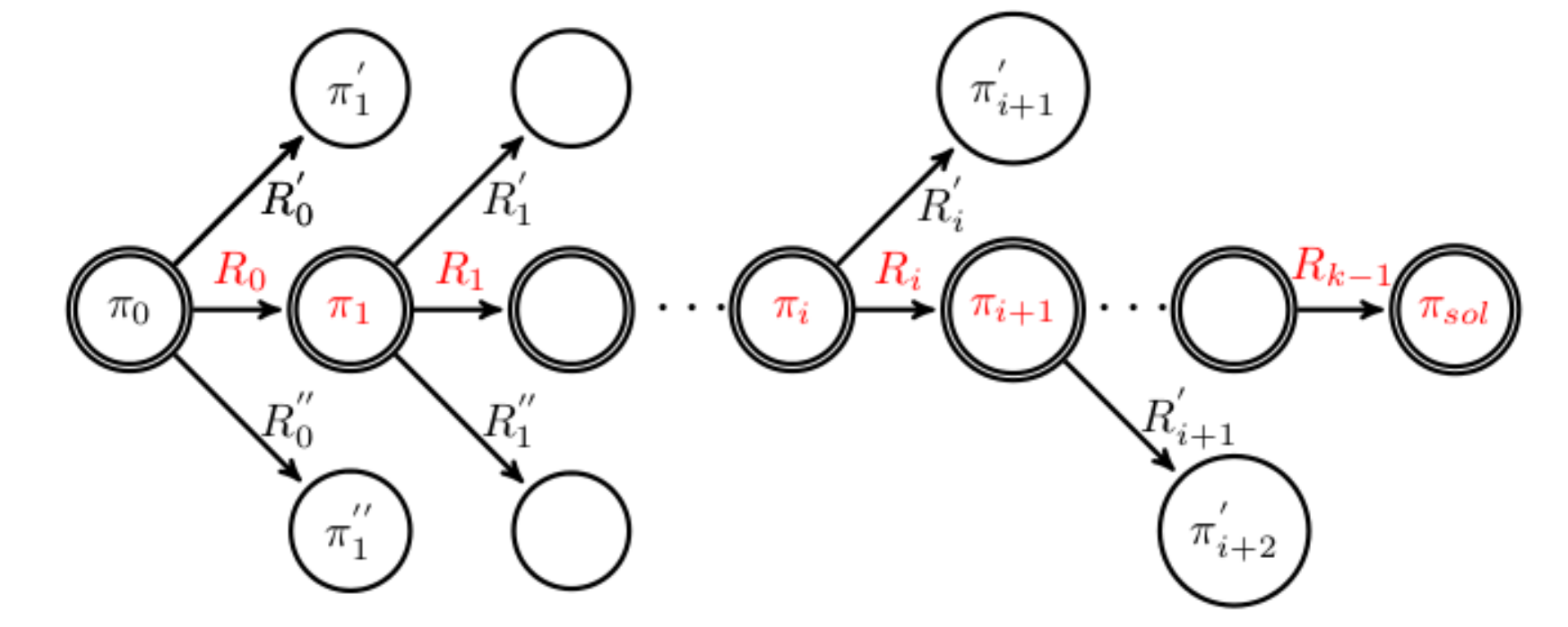}
      \caption{A POCL framework - The refinement \emph{($R$)} starts from $\pi_0$ and it goes to the solution plan $(\pi_{sol})$. 
      At each step, for a selected partial plan, many refinements are possible like refinements of $\pi_0$ which lead to 
      $\pi_1$, $\pi'_1$, and $\pi''_1$. Here, the best child is shown in the horizontal refinements.}
      \label{normal_case}
\end{figure}
The online heuristic adjustment approach is inspired from a recent work presented as technical communication~\cite{ShekharK15}
tested in a few planning domains.
%
We further develop this approach, and provide a complete derivation, and
a proof (in Appendix) of the theorem used in the cited work. 
We assume that
a predictive model $(h)$ for a given partial plan $(\pi)$ 
estimates the actual number of new actions required to be added in $\pi$ to refine it completely to a solution plan,
denoted by $h(\pi)$.
Whilst $h^*(\pi)$ is the minimum number of new actions required for the same purpose~\cite{nguyen2001reviving}. 
%
%
%
In the POCL framework, it is computationally expensive to predict $h^*(\pi)$. 
Since $h$ is also likely to be uninformed, we do adjustments to $h$ by observing 
single-step-error $\mathrm{(\epsilon_h)}$ on-the-fly. 
%
%
%
%
%

%
%
The minimum number of total refinements needed for a partial plan $\pi_i$  to make it a
solution plan, goes through its best child $\pi_{i+1}$ that is obtained after refinement $R_i$. 
A child $\pi_{i+1}$ is the {best} child when it has the lowest prediction of the number of new actions needed 
for its complete refinement among its siblings (Figure~1). We break ties in favor of minimum number of actions in 
the children partial plans.
%
%
%
The set of successors of a partial plan is potentially infinite. 
This is due to the introduction of loops in the plan which simply achieve and destroy subgoals 
(for example: \emph{$\langle$(stack A B), (unstack A B)$\rangle$} or  
\emph{$\langle$(pickup A), (putdown A)$\rangle$}, in Blocksworld domain). 
Such looping is common during refinement process, specially when the heuristic is not well informed. 
We avoid such looping explicitly. This is crucial in real world scenarios. 
For example, a pair of actions like \emph{$\langle$(load-truck~obj~truck~loc), (unload-truck~obj~truck~loc)$\rangle$}, 
in Driverlog domain, could be expensive.
%
%
%
In general in plan space planning there is no backtracking.
Each refinement of a partial plan leads to a different node in the plan space. 
This necessitates that we consider looping explicitly.
Ideally,
\begin{equation}\label{eq1:test}
	h^*(\pi_i) = cost(R_i) + h^*(\pi_{i+1})
\end{equation}
In non-ideal cases where generating $h^*$ is expensive, we assume that,
	$h^*(\pi_i) \approx h(\pi_i) \approx cost(R_i) + h(\pi_{i+1})$.
In Figure~1, for a given partial plan $\pi_i$, 
suppose a minimum of $\mathit{{n}}$ refinement steps are required to make $\pi_i$ a solution plan.
Here we assume that at each stage of the refinement process, the best child for each $\pi_i$ is chosen, 
for all \emph{i}. 
Here, $\mathit{{n}}$ is an estimate of adding new actions in $\pi_i$. 
If we consider only unit cost actions then in the ideal case, $h^*(\pi_{i+1})$ = $h^*(\pi_i)$ - 1. 
Generally, $h$ commits some error at each refinement step called {single-step-error} $\mathrm{(\epsilon_h)}$.
The single-step-error associated with $h$, when $\pi_i$ gets refined by an algorithm using $h$, 
denoted as $\epsilon_{h(\pi_i)}$ is,
\begin{equation}
	\epsilon_{h(\pi_i)} = \big((1 + h(\pi_{i+1})) - h(\pi_{i})\big)
\end{equation}
%
%
%
Using Eq.~\eqref{eq1:test}, we derive (proof is in the Appendix) the following in a non-ideal case: 
\subsubsection{Theorem 1.} 
\emph
{
For a given learned predictive model ($h$) and partial plan ($\pi_i$) in Figure~1 which leads to the solution plan ($\pi_{sol}$)
after certain refinement steps, the enhanced version of the predictive model ($h^e$) is,
\begin{equation}\label{eq2:test}
		h^e(\pi_i) \ = \ h(\pi_{i}) \ + 
				\sum_{	\substack{\pi{'}~\text{{from}} \ \pi_i \rightsquigarrow \pi_{sol}} } \epsilon_{h(\pi')}
\end{equation}
where $\pi_i$ $\rightsquigarrow$ $\pi_{sol}$ is a path in Figure~1 which captures each partial plan ($\pi'$) along 
the path between $\pi_i$ and $\pi_{sol}$.
The path includes $\pi_i$  and excludes $\pi_{sol}$. The term $\epsilon_h$ is single-step-error associated with 
$\mathit{h}$ during refinement.
}
\subsubsection {Enhancement of A Predictive Model}
In Theorem~1, $h^e(\pi_i)$ is an online approximated version of $h^*(\pi_i)$. 
This approximation uses the parent and the best child relationships to measure the step-error of each refinement in the path and 
correct it for the evaluations of further refinements. 
Using Theorem~1 and Figure~1, we calculate the average-step-error associated with $h$, denoted by $\epsilon^{avg}_{h}$,
along the path from $\pi_i$ to $\pi_{sol}$ as,
%
%
\begin{equation}\label{eq31:test}
		\epsilon_{h}^{avg} \ = \ 
					 {\displaystyle \sum_{\substack{\pi'~{\mathit{from}} \ \pi_i \rightsquigarrow \pi_{sol}} 
																		}\epsilon_{h(\pi')}} \Big/ {h^e(\pi_i)}
\end{equation}
Rewriting Eq.~\eqref{eq31:test},
\begin{equation}\label{eq32:test}
{\displaystyle \sum_{\substack{\pi'~{\mathit{from}} \ \pi_i \rightsquigarrow \pi_{sol}}}\epsilon_{h(\pi')}}
	= \epsilon_{h}^{avg} \times {h^e(\pi_i)}
\end{equation}
Using Eq.~\eqref{eq32:test}, Eq.~~\eqref{eq2:test} simplifies to,
\begin{equation}\label{eq33:test}
		h^e(\pi_i) \ = \ h(\pi_{i}) + \epsilon_{h}^{avg} \times {h^e(\pi_i)}			
\end{equation}
Further simplification of Eq.~\eqref{eq33:test} yields,
\begin{eqnarray}\label{eq51:test}
		 h^e(\pi_{i}) =
					\ {h(\pi_{i})} \Big / {(1 - \epsilon_{h}^{avg})} 
\end{eqnarray}  
Another possible expansion, using infinite geometric progression, of Eq.~\eqref{eq51:test} would be,
\begin{eqnarray}\label{eq51:test}
		 h^e(\pi_{i}) = \ {h(\pi_{i})} \times \sum_{i=0}^{\infty}{(\epsilon_{h}^{avg})}^i
\end{eqnarray}  
We use RegPOCL to test the effectiveness of $h^e(\pi_{i})$ in the POCL framework, where it selects the best partial plan.
\section{Experiment Design}
In this section we describe the evaluation phase settings. 
This includes
\begin{inparaenum}[(\itshape i\upshape)]
\item the heuristics selected as features, and
\item the domains selected.
\end{inparaenum}
\subsection{Feature Heuristics for Learning} The features used for learning are non-temporal heuristics from the literature of 
POCL planning. 
Considering the applicability of some of the POCL heuristics in the literature~\cite{younes2003vhpop,nguyen2001reviving}, 
we select six different heuristic functions. 
Some of these heuristics are informed but their informativeness varies over different planning domains. 
%
Our aim is to learn a more informed combination from these individual heuristics. 
The six heuristics are,
\subsubsection{G Value ($h_{{g\text{-}val}}$)} 
This returns the number of actions in a selected partial plan 
$\pi$ not counting the two dummy actions ($a_0$ and $a_{\infty}$). 
It signifies how far the search has progressed from the starting state.
\subsubsection{Number of Open Conditions ($h_{{OC}}$)} 
This is total number of unsupported causal links present in a partial plan,
$h_{{OC}}(\pi) = \left| \mathrm{OC} \right|$~\cite{nguyen2001reviving}.
\subsubsection{Additive Heuristic ($h_{{add}}$)} 
The additive heuristic $h_{{add}}$~\cite{haslum2000admissible}, 
adds up the steps required by each individual open goal. 
Younes and Simmons (2003) use an adapted version of additive heuristic in POCL planning for the first time.
\subsubsection{Additive Heuristic with Effort ($h_{{add,w}}$)}  
The estimate is similar to $h_{add}$ but it considers the cost of an action as the 
number of preconditions of that action, plus the {linking cost} \textbf{1} if the action supports any unsupported causal 
link~\cite{younes2003vhpop}. We call it $h_{{add,w}}$ as its notation is not used earlier. Here, $w$ signifies the extra work
required.
\subsubsection{Accounting for Positive Interaction ($h_{{add}}^{r}$)}
This returns an estimate which takes into account the positive interactions between subgoals 
while ignoring the negative interactions. 
This is represented as $h_{{add}}^{r}$ that is a variant of $h_{add}$~\cite{younes2003vhpop}. 

\subsubsection{Accounting for Positive Interaction with Effort ($h_{{add,w}}^{r}$)}
This is similar to the above heuristic which considers the total effort required~\cite{younes2003vhpop}. 
A standard notation of this heuristic is also not used in the literature.
%
%
%
\subsection{Domains Selected} 
%
%
We consider the following domains: 
%
Logistics and Gripper from IPC~1, Logistics and Elevator from IPC~2, Rovers and Zenotravel from IPC~3,
and Rovers from IPC~5. 
In our experiments we do not consider other domains from these competitions because either the state-of-the-art 
heuristics are not able
to create enough training instances for learning, or {RegPOCL} does not support the domain definition language features. 
IPC~4 domains are not selected since the planner is not able to generate 
enough instances to initiate offline learning. 
The domains from IPC~6 and later are not supported by {RegPOCL} 
because the representations use action costs, fluents, and hard and soft constraints.
Some of them can be included by some preprocessing like removal of actions with cost from the domain description files.
%
%
%
%

%
For each selected domain, we consider problems that are represented in STRIPS style. 
We select small sized problems for learning and test the learned predictive models over large sized problems in the same domain. 
We have a total of 109 small sized problems from the selected domains. 
The last four feature heuristics from the previous subsection have been used for calculating targets in each domain. 
This means that we generate four different datasets in each selected domain from which best two are selected.
We choose satisficing track problems for generating training instances.
For the training set preparation, we fix a time limit of 3 minutes and 
an upper limit of 500,000 on the node generation.
We generate a few thousand training instances except for the Zenotravel domain where the total instances are 950. 
To avoid overfitting, we pick training instances between 250 to 350 from the larger training sets.
%
%
%
%
%
%
\section{Selected Learning Approaches}
In this section, we discuss in brief a procedure for feature selection in each dataset for 
training regression models,
and different regression techniques with their references.
\subsection{Feature Selection} 
In general, the training sets contain irrelevant or redundant attributes
(out of the six selected heuristics).
To reduce the training effort and increase the efficiency of our planner,
we discard them from the training set.
The planner is bound to calculate all the selected features at each stage of refinement. 
The correlation based feature selection technique~\cite{hall1999correlation} is used to find the correlated features.
%
%
%
%
\begin{table*}[t]
\centering
\scalebox{0.70}
{ 
\begin{tabular}
{ @{} l | r || r r | r r || r r | r r || r r r r || r r r r ||r @{}}
\hline
	\multirow{3}{*}{\textbf{Domain}} &
	\multirow{3}{*}{\textbf{\#}} &
	\multicolumn{8}{c||}{\textbf{POCL Heuristics}} &
	\multicolumn{8}{c||}{\textbf{Evaluation using Fast Downward (FD)}} &
	\multirow{3}{*}{\textbf{LAMA}} 
	\\ \cline{3-18}
	&
	&
	\multicolumn{4}{c||}{\textbf{State-of-the-art}} &
	\multicolumn{4}{c||}{\textbf{Via learning approaches}} &
	\multicolumn{4}{c||}{\textbf{Lazy}} &
	\multicolumn{4}{c||}{\textbf{Eager}} &
	\\ \cline{3-18} 
	 & &
	\multicolumn{1}{c}{$h_{{add}}$} & 
	\multicolumn{1}{c|}{$h_{{add,w}}$} & 
	\multicolumn{1}{c}{$h_{{add}}^r$} & 
	\multicolumn{1}{c||}{$h_{{add,w}}^{r}$} & 
	\multicolumn{1}{c}{$h_{{add}}^{{l}}$} & 
	\multicolumn{1}{c|}{$h_{{add}}^{l,e}$} & 
	\multicolumn{1}{c}{$h_{{add,w}}^{{l}}$} & 
	\multicolumn{1}{c||}{$h_{{add,w}}^{{l,e}}$} &
	\multicolumn{1}{c}{FF} & 
	\multicolumn{1}{c}{CEA} &
	\multicolumn{1}{c}{LM-Cut} &
	\multicolumn{1}{c||}{MHFS} &
	\multicolumn{1}{c}{FF} & 
	\multicolumn{1}{c}{CEA} &
	\multicolumn{1}{c}{LM-Cut} &
	\multicolumn{1}{c||}{MHFS} &	
	\\ 	\hline \hline
	Gripper-1 	&20		&16		&\textbf{20}		&1		&1		&\textbf{20}		&\textbf{20}	&\textbf{20}		&\textbf{20}		
						&\textbf{20} 	&\textbf{20}		&\textbf{20} 	&\textbf{20}		
						&\textbf{20} 	&\textbf{20}		&\textbf{20} 	&\textbf{20}		&\textbf{20} \\
	Rovers-3 	&20		&19		&19		&\textbf{20}		&\textbf{20}		&\textbf{20}		&\textbf{20}		&\textbf{20}		
																														&\textbf{20}																	
						&18		&17		&14		&\textbf{20}		&19		&18		&15		&\textbf{20}		&\textbf{20}	 \\
	Rovers-5	&40		&28		&31		&32		&36		&${}^{\text{\small +}}$39		&${}^{\text{\small +}}$39	
						&${}^{\text{\small +}}$36		&${}^{\text{\small +}}$39															
						&22		&25		&15		&28		&24		&29		&17		&30		&\textbf{40} \\
	\cline {1-19} 			 
    &
	&
	\multicolumn{1}{c}{$h_{{add}}$} & 
	\multicolumn{1}{c|}{$h_{{add,w}}$} & 
	\multicolumn{1}{c}{$h_{{add}}^r$} & 
	\multicolumn{1}{c||}{$h_{{add,w}}^{r}$} & 
	\multicolumn{1}{c}{$h_{{add}}^{{r,l}}$} & 
	\multicolumn{1}{c|}{$h_{{add}}^{r,l,e}$} & 
	\multicolumn{1}{c}{$h_{{add,w}}^{{r,l}}$} & 
	\multicolumn{1}{c||}{$h_{{add,w}}^{{r,l,e}}$} &
	\multicolumn{1}{c}{FF} & 
	\multicolumn{1}{c}{CEA} &
	\multicolumn{1}{c}{LM-Cut} &
	\multicolumn{1}{c||}{MHFS} &
	\multicolumn{1}{c}{FF} & 
	\multicolumn{1}{c}{CEA} &
	\multicolumn{1}{c}{LM-Cut} &
	\multicolumn{1}{c||}{MHFS} & \\ 
	%
	\hline
	Logistics-1	&35 	&25		&1		&32		&28		&${}^{\text{\small +}}$32		&${}^{\text{\small +}}$32 	&22
							&${}^{\text{\small +}}$32	
							&25		&34		&21		&28		&28		&34		&16		&25		&\textbf{35} \\
	Elevator-2		&150	&148	&14		&\textbf{150}	&58		&\textbf{150}	&\textbf{150}	&\textbf{150}	&\textbf{150}										
							&\textbf{150}	&\textbf{150}	&\textbf{150}	&\textbf{150}
							&\textbf{150}	&\textbf{150}	&\textbf{150}	&\textbf{150}	&\textbf{150}	\\
	Logistics-2	&40		&36		&12		&36		&34		&\textbf{40}	&\textbf{40}	&\textbf{40}	&\textbf{40}
							&36		&36		&35		&36		&36		&36		&36		&36		&\textbf{40}	\\
	Zenotravel-3	&20		&5		&4		&9		&10		&${}^{\text{\small +}}$16		&${}^{\text{\small +}}$16
							&${}^{\text{\small +}}$16		&${}^{\text{\small +}}$16
							&\textbf{20}		&\textbf{20} 	&17		&\textbf{20}	
							&\textbf{20}		&\textbf{20} 	&16		&\textbf{20}	&\textbf{20} \\
\hline			 
\end{tabular}
}
\caption
{
Number of solved problems using each heuristic. 
\textbf{\#} is the number of problems in each domain.
State-of-the-art POCL heuristics are compared with the learned ones in the left half. 
The state-based heuristics FF, CEA, and LM-Cut 
and their combination using MHFS strategy are also compared with POCL heuristics.
The last column captures the performance of LAMA11. 
Best results are shown in \textbf{bold}, 
and a number with ``+'' mark (\emph{e.g.}${}^{\text{\small +}}$36) shows competitive performance by 
the learned models and their enhancements over each base heuristic. 
Similar representations are followed in other tables.
}
\end{table*}
%
\subsection{Regression Techniques}
We use the following regression techniques to learn predictive models. These techniques have been applied in planning 
for learning in recent years~\cite{samadi2008learning,thayer2011learning,us2013learning}.
\subsubsection{Linear Regression (LR)}
The regression model learns a linear function that minimizes the sum of squared 
error over the training instances~\cite{bishop2006pattern}. 

\subsubsection{M5P}
M5P gives more flexibility than LR due to its nature of 
capturing non linear relationships.
M5P technique learns a regression tree~\cite{quinlan1992learning} that approximates the class value. 

\subsubsection{M5Rules}
Similar to M5P but generates rules instead of modeling regression trees~\cite{quinlan1992learning}.

\subsubsection{Least Median Squared (LMS)}
LMS is similar to LR with median squared error.
Functions are generated from subsamples of data with least squared error function. 
Usually a model with lowest median squared error is selected~\cite{rousseeuw2005robust}.

\subsubsection{Multilayer Perceptron (MLP)} 
MLP can learn more complex relationships compared to the other four 
regression techniques~\cite{bishop2006pattern}.

The techniques discussed above are used to learn models through WEKA~\cite{Hall2009} using 
a 10-fold cross-validation in each domain. 
A regression technique called SVMreg that implements support vector machine for regression purposes, 
is not used in this work due to some technical difficulty. 
However, it has not much influenced planning processes in the past~\cite{us2013learning}.
%
%
\section{Experimental Evaluation} 
We use MC-Loc and MW-Loc~\cite{younes2003vhpop} as flaw selecting heuristics for refining a partial plan.
%
They give higher preference to the local flaws present in the partial plan.  
We employ Greedy Best First Search algorithm for selecting the next partial plan for refinement.
\subsection{Environment}
We perform the experiments on Intel Core~2 Quad with 2.83~GHz 64-bit processor and 4GB of RAM. 
To evaluate the effectiveness of learned models and to correct the single-step-error associated with the models, 
a time limit of 15 minutes and a node generation limit of 1 million is used.
\subsection{Evaluations}
We use RegPOCL to compare the performances of the offline predictive models $h^l$, their corresponding enhanced models
$h^{l,e}$ and the last four base features. 
%
%
%
These are also compared with some of the recent effective state-space based heuristics and approaches that are introduced later.
%
%
%
We exclude the first two base features from comparison since they are weak heuristics and RegPOCL 
does not solve sufficient problems using them.
However, they are useful while working jointly with other informed heuristics. 
Next, we discuss the observations made during the training phase. 
\subsubsection{Training} 
Using Algorithm~1, we prepare datasets and learn different predictive models by applying the various regression techniques 
discussed earlier. 
We select each of the last four features to solve a set of problems.
The target value is the plan length found by RegPOCL using the base features.  
The dataset preparation phase took less than two hours on an average in each domain, for each of the four features.
Once we have enough instances, we begin the training process. 
In each domain, the best two datasets out of four with sufficient training points
is selected for model training by visualizing the distribution of training points using WEKA.   
Note that, in Figure~1, different heuristics for calculating target values will prefer different paths. 
Therefore, the four base features will generate four different datasets. 
The attribute selection strategy allows us to select a subset of attributes in the training set by removing
correlated and redundant features.
We learn a total of 70 (7 domains $\times$ 2 datasets $\times$ 5 regression techniques) regression models. 
The training phase took 20ms (milliseconds) using LR, 270ms using LMS, 600ms using MLP, 82ms  
using M5Rule, and  58ms using M5P on an average per model.  
All the learned models have high accuracy but LR is the fastest, followed by M5P and M5Rule.
Next, we test these models on different benchmarks.
\subsubsection{Testing}
We test the effectiveness of our approaches by selecting partial plans $(\pi)$ for refinement using RegPOCL. 
We assume that an informed heuristic leads to minimal possible refinements needed for $\pi$.
Next, for the comparison we compute score\footnote{\url{https://helios.hud.ac.uk/scommv/IPC-14/}} 
as in IPC for satisficing track problems.
The better score on each standard signifies the better performance.
We compare the performance of the learned models with the selected base features 
$h_{add}$, $h_{add,w}$, $h_{add}^{r}$, and $h_{add,w}^{r}$. 
The comparison is done on the basis of 
\begin{inparaenum}[(\itshape i.\upshape)]
\item the number of solved problems, and 
\item the score obtained on plan quality, execution time, nodes (partial plan) visited, and makespan quality.
\end{inparaenum}
For example, the offline learned model $h_{add}^{r,l}$ is learned on a dataset prepared using $h_{add}^{r}$. 
In other words, RegPOCL uses $h_{add}^{r}$ for calculating the target values in the dataset.
$h_{add}^{r,l}$ can be enhanced to $h_{add}^{r,l,e}$ 
using the online heuristic adjustment approach which is expected to be more informed than $h_{add}^{r,l}$. 
It is similar for other learned heuristics.
These models are applied in the POCL framework for selecting the most suitable partial plan,
followed by the heuristic MW-Loc~\cite{younes2003vhpop} for selecting the most adverse flaw in it .

We also compare our approaches with state-space based approaches on the basis of the number of problems solved, and 
score obtained on plan quality and total execution time. 
We select fast forward heuristic (FF)~\cite{hoffmann2001ff}, context-enhanced additive heuristic (CEA)~\cite{helmert2008unifying},
and landmark-cut heuristic (LM-Cut)~\cite{helmert2011lm}. 
We also use these heuristics together by applying them in multi-heuristic first solution strategy (MHFS)~\cite{roger2010more}. 
In general, the strategy performs better with alternating usage of different heuristics instead of combining them. 
We also compare the effectiveness of our techniques with 
LAMA11~\cite{richter2011lama}; the winner of IPC-2011 in the sequential satisficing track. 
LAMA11 applies FF and LM-Count~\cite{richter2008landmarks} heuristics together using multi-queue search.
We set a 20 minute time limit while evaluating LAMA11 over these domains,
since it has an internal time limit of 5 minutes for the invariant synthesis part of translator.
%
%
%
%
\begin{table*}[t]
\centering
\scalebox{0.7}
{ 
\begin{tabular}
{ @{} l || r r | r r || r r | r r || r r r r || r r r r ||r @{}}
\hline
	\multirow{3}{*}{\textbf{Domain}} &
	\multicolumn{8}{c||}{\textbf{POCL Heuristics}} &
	\multicolumn{8}{c||}{\textbf{Evaluation using Fast Downward (FD)}} &
	\multirow{3}{*}{\textbf{LAMA}} 
	\\ \cline{2-17}
	&
	\multicolumn{4}{c||}{\textbf{State-of-the-art}} &
	\multicolumn{4}{c||}{\textbf{Via learning approaches}} &
	\multicolumn{4}{c||}{\textbf{Lazy}} &
	\multicolumn{4}{c||}{\textbf{Eager}} & 
	\\ \cline{2-17} 
	&
	\multicolumn{1}{c}{$h_{{add}}$} & 
	\multicolumn{1}{c|}{$h_{{add,w}}$} & 
	\multicolumn{1}{c}{$h_{{add}}^r$} & 
	\multicolumn{1}{c||}{$h_{{add,w}}^{r}$} & 
	\multicolumn{1}{c}{$h_{{add}}^{{l}}$} & 
	\multicolumn{1}{c|}{$h_{{add}}^{l,e}$} & 
	\multicolumn{1}{c}{$h_{{add,w}}^{{l}}$} & 
	\multicolumn{1}{c||}{$h_{{add,w}}^{{l,e}}$} &
	\multicolumn{1}{c}{FF} & 
	\multicolumn{1}{c}{CEA} &
	\multicolumn{1}{c}{LM-Cut} &
	\multicolumn{1}{c||}{MHFS} &
	\multicolumn{1}{c}{FF} & 
	\multicolumn{1}{c}{CEA} &
	\multicolumn{1}{c}{LM-Cut} &
	\multicolumn{1}{c||}{MHFS} &	
	\\ 	\hline \hline
	Gripper-1 	&16.0	&14.9	&0.7	&0.7		&\textbf{20.0}		&\textbf{20.0}		&\textbf{20.0}		&\textbf{20.0}		
						&15.45 		&14.9			&15.5 		&15.5		
						&15.5		&14.9			&15.5 	&15.5	&\textbf{20.0} \\
	Rovers-3 	&17.3	&16.2	&18.1	&16.9		&17.8		&${}^{\text{\small +}}$18.6	&17.8 &${}^{\text{\small +}}$18.6
						&17.1	&15.9	&12.5	&18.8		&18.1		&16.9		&13.9	&19.0		&\textbf{19.8}	 \\
	Rovers-5	&25.7		&26.3		&28.0	&30.2		&${}^{\text{\small +}}$33.1	&${}^{\text{\small +}}$36.3	&30.1		
						&${}^{\text{\small +}}$33.6	&20.8		&23.4	&13.5		&26.0		&22.7	&27.3	&15.7	&28.3		
						&\textbf{39.8} \\
%
\hline
    &
	\multicolumn{1}{c}{$h_{{add}}$} & 
	\multicolumn{1}{c|}{$h_{{add,w}}$} & 
	\multicolumn{1}{c}{$h_{{add}}^r$} & 
	\multicolumn{1}{c||}{$h_{{add,w}}^{r}$} & 
	\multicolumn{1}{c}{$h_{{add}}^{{r,l}}$} & 
	\multicolumn{1}{c|}{$h_{{add}}^{r,l,e}$} & 
	\multicolumn{1}{c}{$h_{{add,w}}^{{r,l}}$} & 
	\multicolumn{1}{c||}{$h_{{add,w}}^{{r,l,e}}$} &
	\multicolumn{1}{c}{FF} & 
	\multicolumn{1}{c}{CEA} &
	\multicolumn{1}{c}{LM-Cut} &
	\multicolumn{1}{c||}{MHFS} &
	\multicolumn{1}{c}{FF} & 
	\multicolumn{1}{c}{CEA} &
	\multicolumn{1}{c}{LM-Cut} &
	\multicolumn{1}{c||}{MHFS} & \\ 
	\hline
	%
	Logistics-1	&24.3	&0.8	&31.2	&26.1	
							&${}^{\text{\small +}}$31.2	&${}^{\text{\small +}}$31.2	&20.9		&\textbf{32.0}
							&23.7		&30.3		&19.7		&27.4		&27.0			&30.8		&15.1		&24.6		&30.8 \\
	Elevator-2		&142.8	&11.8	&144.9	&49.7		&142.7	
							&${}^{\text{\small +}}$144.9		&142.7
							&${}^{\text{\small +}}$144.9										
							&117.1	&112.0	&117.1	&116.9
							&144.1	&136.0	&144.2	&141.0	&\textbf{148.2}	\\
	Logistics-2	&33.8		&10.7		&34.9		&30.7		&\textbf{38.1}		&${}^{\text{\small +}}$36.3	
																			&\textbf{38.1}		&${}^{\text{\small +}}$36.3
							&33.72		&29.5		&32.4		&33.3		&34.1		&30		&33.7		&35.6			&37.8	\\
	Zenotravel-3	&4.6		&3.7		&8.5		&8.8		&${}^{\text{\small +}}$13.5	&${}^{\text{\small +}}$13.7	
							&${}^{\text{\small +}}$13.5 	&${}^{\text{\small +}}$13.7
							&17.3 	&17.7		&14.7 	&18.6	
							&18.1		&14.3 	&16		&18.8	&\textbf{19.2} \\
	\hline \hline
	\multicolumn{1}{@{}c||}{\textbf{Time Score}}
							&204.4		&76.5		&197.7	 	&148.6		&${}^{\text{\small +}}$272.9		&${}^{\text{\small +}}$260.8
							&${}^{\text{\small +}}$258.6		&${}^{\text{\small +}}$264.3		&255.6	
							&\textbf{280.0}		&228.0		&233.8		&259.6		&271.3		&223.0		&222.5		&137 \\					
\hline			 
\end{tabular}
}
\caption
{
Scores on plan quality and overall time. We compare state-of-the-art POCL heuristics with learned ones. 
The effectiveness of the POCL heuristics is compared with some latest state based approaches. 
The last row demonstrates the overall time score of each heuristic. 
The numerical representations and column details are similar to Table~1.
}
\end{table*}
All the state-based approaches are evaluated using Greedy Best First Search algorithm in 
the fast downward planning system~\cite{helmert2006fast}.
We use ``eager'' and ``greedy'' types of evaluations with no preferred operators.
%
%
%
%
The regression models selected in the chosen domains are trained using,
\begin{inparaenum}[(\itshape i\upshape)]
\item M5P in Gripper-1 and Elevator-2, 
\item LR in Rovers-3, Rovers-5, Logistics-2, and Zenotravel-3, and
\item M5Rule in Logistics-1.
\end{inparaenum}
In Table~1, we compare 
\begin{inparaenum}[(\itshape i\upshape)]
\item the base features, 
\item offline learned models and their enhanced versions
\item state-space based heuristics FF, CEA, and LM-Cut, and 
\item the strategies used in MHFS, and LAMA11, 
\end{inparaenum}
on the basis of number of problems solved. 
In this table, RegPOCL solves equal number of problems as LAMA11
in Gripper-1, Rovers-3, Elevator-2, and Logistics-2 using each of learned heuristics and their enhanced versions.
The base features have performed well in some domains but are not consistent overall. 
In Rovers-5, our approaches solved 1 problem less than LAMA11, 
but they beat other state-space based competitors comprehensively. 
Also, each learned model has performed better than all the base features. 
For Logistics-2, we are competitive with LAMA11 and  solve at least 4 more problems than other good heuristics 
like CEA and LM-Cut.
In Zenotravel-3, RegPOCL solved 6 problems more by applying our approaches but loses to the state-based competitors.
Our second approach improves the performance of the learned models in Rovers-5 by solving 3 more problems, 
and in Logistics-1 where it solves 10 more problems.  
This approach could not increase the coverage in other domains. 
LAMA11 wins on the basis of the total number of problems solved in each domain.

In Table~2, we compare the score obtained on plan quality by each of the 
base features, 
learned models with their corresponding enhancements, and
state-space based heuristics and techniques. 
%
%
\begin{table}[t]
\centering
\scalebox{0.65}
{ 
\begin{tabular}
{@{} l || r r | r r || r r | r r @{}}
\hline
%
%
	\multirow{2}{*}{\textbf{Domain}} &
	\multicolumn{4}{c||}{\textbf{State-of-the-art}} &
	\multicolumn{4}{c}{\textbf{Via learning approaches}} 
	\\ \cline{2-9} 
	&
	\multicolumn{1}{c}{$h_{{add}}$} & 
	\multicolumn{1}{c|}{$h_{{add,w}}$} & 
	\multicolumn{1}{c}{$h_{{add}}^r$} & 
	\multicolumn{1}{c||}{$h_{{add,w}}^{r}$} & 
	\multicolumn{1}{c}{$h_{{add}}^{{l}}$} & 
	\multicolumn{1}{c|}{$h_{{add}}^{l,e}$} & 
	\multicolumn{1}{c}{$h_{{add,w}}^{{l}}$} & 
	\multicolumn{1}{c@{}}{$h_{{add,w}}^{{l,e}}$} 
	\\ 	\hline \hline
	Gripper-1 	&7.0	&14.0	&0.0	&0.0	&${}^{\text{\small +}}$18.0  &\textbf{19.7}	&${}^{\text{\small +}}$18.0	&\textbf{19.7} \\
	Rovers-3 	&8.6	&8.3	&12.6 	&\textbf{19.0}		&13.3	&13.6 	&13.3	&13.6 \\
	Rovers-5	&8.2	&14.7	&13.5	&30.9		&\textbf{31.4}		&\textbf{31.4}		&24.7 &\textbf{31.4}	\\
%
	\hline
    &
	\multicolumn{1}{c}{$h_{{add}}$} & 
	\multicolumn{1}{c|}{$h_{{add,w}}$} & 
	\multicolumn{1}{c}{$h_{{add}}^r$} & 
	\multicolumn{1}{c||}{$h_{{add,w}}^{r}$} & 
	\multicolumn{1}{c}{$h_{{add}}^{{r,l}}$} & 
	\multicolumn{1}{c|}{$h_{{add}}^{r,l,e}$} & 
	\multicolumn{1}{c}{$h_{{add,w}}^{{r,l}}$} & 
	\multicolumn{1}{c}{$h_{{add,w}}^{{r,l,e}}$} \\	
	\hline
	Logistics-1	&5.1	&0.6	&16.5	&21.6	
							&\textbf{29.9}		&\textbf{29.9}		&{20.7}	&\textbf{29.9} \\
	Elevator-2		&41.4	&8.2	&58.9	&39.7		&\textbf{145.0} 		
														&${}^{\text{\small +}}$135.0	&\textbf{145.0}	&${}^{\text{\small +}}$135.0	\\
	Logistics-2	&24.5	&6.5	&23.4	&26.7		&${}^{\text{\small +}}$33.4	
														&\textbf{35.3}	 &${}^{\text{\small +}}$33.4	&\textbf{35.3}   \\
	Zenotravel-3	&0.0	&1.0	&2.4	&7.7 	&7.4	&\textbf{11.07}	&7.4	&\textbf{11.07} \\
	\hline 
\end{tabular}
}
\caption
{
Results of refinement score (the number nodes visited) of RegPOCL using each heuristic. 
We compare state-of-the-art heuristics with the learned models ($h^{l}$) and their further enhancements ($h^{l,e}$). 
The best results are in \textbf{bold}.
}
\end{table}
%
%
\begin{table}[t]
\centering
\scalebox{0.64}
{ 
\begin{tabular}
{@{}l || r r | r r || r r | r r@{}}
\hline
%
%
	%
	\multirow{1}{*}{\textbf{Domain}} &
	\multicolumn{1}{c}{$h_{{add}}$} & 
	\multicolumn{1}{c|}{$h_{{add,w}}$} & 
	\multicolumn{1}{c}{$h_{{add}}^r$} & 
	\multicolumn{1}{c||}{$h_{{add,w}}^{r}$} & 
	\multicolumn{1}{c}{$h_{{add}}^{{l}}$} & 
	\multicolumn{1}{c|}{$h_{{add}}^{l,e}$} & 
	\multicolumn{1}{c}{$h_{{add,w}}^{{l}}$} & 
	\multicolumn{1}{c@{}}{$h_{{add,w}}^{{l,e}}$} 
	\\ 	\hline \hline
	Gripper-1 	&16.0	&9.8	&0.5	&0.5		&\textbf{20.0}  &\textbf{20.0}	&\textbf{20.0}	&\textbf{20.0} \\
	Rovers-3 	&\textbf{17.9}	&15.5	&17.8	&15.8		&17.4		&{17.7} 	&17.4		&{17.7} \\
	Rovers-5	&26.1	&24.2	&28.4	&26.5		&\textbf{30.5}		&${}^{\text{\small +}}$29.8	
																							&${}^{\text{\small +}}$27.6	&\textbf{30.5} \\
	%
	\hline
    &
	\multicolumn{1}{c}{$h_{{add}}$} & 
	\multicolumn{1}{c|}{$h_{{add,w}}$} & 
	\multicolumn{1}{c}{$h_{{add}}^r$} & 
	\multicolumn{1}{c||}{$h_{{add,w}}^{r}$} & 
	\multicolumn{1}{c}{$h_{{add}}^{{r,l}}$} & 
	\multicolumn{1}{c|}{$h_{{add}}^{r,l,e}$} & 
	\multicolumn{1}{c}{$h_{{add,w}}^{{r,l}}$} & 
	\multicolumn{1}{c}{$h_{{add,w}}^{{r,l,e}}$}  \\	
	\hline
	Logistics-1	&21.8	&0.9	&25.2	&20.8	
							&${}^{\text{\small +}}$30.4	&${}^{\text{\small +}}$30.4	&17.6	&\textbf{30.5} \\
	Elevator-2		&147.1	&11.7	&\textbf{149.1}	&50.5		&	\textbf{149.1}	
							&146.8		&\textbf{149.1} &	146.8	\\
	Logistics-2	&26.3	&7.0	&33.0	&19.3	&${}^{\text{\small +}}$33.5	&\textbf{35.4}	 
							&${}^{\text{\small +}}$33.4	&\textbf{35.4}   \\
	Zenotravel-3	&3.7	&2.7	&7.5	&7.7		&${}^{\text{\small +}}$12.1
									&\textbf{13.5}		&${}^{\text{\small +}}$12.1		&\textbf{13.5} \\
	\hline 
\end{tabular}
}
\caption
{
Makespan quality score of the heuristics. Columns are identical to corresponding ones in Table~3.
}
\end{table}
LAMA11 is an anytime planner which gives solution close to the optimal but takes more time to compute 
shorter plans.
For Gripper-1 and Logistics-2, RegPOCL using the learned heuristics solves equal number of problems but finds shorter 
plans compared to LAMA11. 
In Logistics-1, RegPOCL lost to LAMA11 by 3 problems in the number of problems solved, 
but obtained a higher score 
in the other problems it solved as it produces shorter plans 
using $h_{add,w}^{r,l,e}$, $h_{add}^{r,l}$, and $h_{add}^{r,l,e}$.
The effectiveness of $h_{add,w}^{r,l,e}$ wins in this domain with best plan quality. 
$h_{add,w}^{r,l,e}$ also increases the plan quality score from 20.9 that is obtained by $h_{add,w}^{r,l}$ to 32.0
using our online error correcting strategy. 
However, in Logistics-2, the performance has decreased after further enhancements 
of $h_{add}^{r,l}$ and $h_{add,w}^{r,l}$.

In general, the performance of RegPOCL using either of the offline learned models and their enhancements 
is often better than that using the base features.
In most of the cases, the online error adjustment approach has further enhanced the performance of these learned models. 
The last row of Table~2 gives the score obtained on total time taken by the process. 
If a planner takes less than 1 second for solving a problem then it gets full score. 
On the total time score the winner is CEA with ``lazy'' evaluation.
The learned models and their enhanced versions 
have obtained better scores than other competitors except CEA. 
These scores are very close to the winning score and almost twice that of LAMA11. 

In Table~3, we compare the score obtained on the number of nodes RegPOCL visits for solving the planning problems.
This is obtained for the base features, the learned heuristics, and their enhanced versions. 
The models obtained after offline learning are more informed toward goals and refines fewer partial plans. 
The score obtained by the learned models is further increased by a good factor in Zenotravel-3 
using the error correcting approach. 
For Elevator-2, the error correcting approach has shown some negative effect which continues in Table~4 too. 
In Table~4, we demonstrate the score obtained on the makespan quality. 
Higher score signifies smaller makespan and more flexibility in the generated solution plan.
In Elevator-2 and Rovers-5, 
the scores of $h_{add}^l$ and $h_{add}^{r,l}$ have decreased due to the negative effects of our error adjustment approach,
while the score obtained by $h_{add,w}^{r,l,e}$ is almost 1.5 times the score of $h_{add,w}^{r,l}$ in Logistics-1. 
In general, the offline learned models have generated more flexible plans with 
shorter makespan than the base features. 
These qualities are further improved using the enhanced versions of these models.
\section{Discussion}
We have already discussed the advantages of our approaches but they also have limitations. 
In our offline approach,
we are bound to do some poor generalization while learning heuristics.
%
%
%
Current literature supports the idea of selecting a large feature set for more accurate learning~\cite{RobertsHWd08}. 
Accuracy can also be improved using an empirical performance model of all components of a portfolio to decide 
which component to pick next~\cite{FawcettVH0HL14}.
In our work, a large feature set may have some drawbacks. For example, computing the features at each refinement step during 
the planning process is computationally expensive.
The online error adjustment approach could also perform poorly in certain domains. 
In Figure~1, if the orientation of objects in a domain is such that $h(\pi_{i+1})$ is larger than $h(\pi_{i})$
then $\epsilon_{h({\pi_{i})}}$ may not be accurate. The inaccuracy in $\epsilon_{h({\pi_{i})}}$ is compounded 
if the above condition holds at the beginning of the planning process.
This results in an inaccurate $\epsilon_{h}^{avg}$ value, leading to wrong selection of the partial plan to refine next. 
Consequently, the planner may end up finding longer and less flexible plans. 
Another limitation is that 
a refinement may change the existing priorities of partial plans in the set due to the single-step-error adjustment.
Considering the time factor, we avoid changing the decided priorities of those partial plans.
This may also lead to inaccuracy in $\epsilon_{h}^{avg}$.

%
%
%

Our approaches do not utilize the advantage of strategies like alternation queue, and 
candidate selection using concept of pareto optimality~\cite{roger2010more}. 
%
%
%
Recently, the planning community has tried coming up with effective portfolios of heuristics or planners. 
The techniques of generating good portfolios are not new to theoretical machine learning.
A follow up work done in the past is 
combining multiple heuristics online~\cite{StreeterGS07}. 
One could form a portfolio of different algorithms to reduce the total makespan for a set of jobs 
to solve~\cite{StreeterS08}.
The authors provide a bound on the performance of the portfolio approaches. 
For example, an execution of a greedy schedule of algorithms cannot exceed four times the optimal schedule. 
%
%
%

%
In planning, a sequential portfolio of planners or heuristics aims to optimize the performance metrics. 
In general, such configurations automatically generate sequential orderings of best planning algorithms. 
In the portfolio the participants are allotted some timestamp to participate in solving problems in the ordering.
A similar approach is used in ~\cite{SeippSHH15}. 
The authors outline their procedure for optimal and satisficing planning.
The procedure used in this work starts with a set of planning algorithms and a time bound.
It uses another procedure \emph{OPTIMIZE} that focuses on the marginal improvements of the performance.
Here, the quality of the portfolio is bounded by $\mathrm{(1-(1/e))~\times}$~\sc {opt}, \normalfont and 
the running time cannot exceed {4}~\sc{opt}. \normalfont
The components can be allowed to act in a round-robin fashion~\cite{gerevini2014planning}.
\textcolor{black}
{
The state-of-the-art planners exhibit variations in their runtime for a given problem instance,
so no planner always dominates over others. 
A good approach would be to select a planner for a given instance by looking at its processing time.
This is done by building an empirical performance model (EPM) for each planner.
EPM is derived from sets of planning problems and performance observation.
It predicts whether the planner could solve a given instance~\cite{FawcettVH0HL14}.  
The authors consider a large set of instance features and 
show that the runtime predictor is often superior to the individual planners. 
}
Performance wise sorting of components in a portfolio is also possible~\cite{NunezBL15}.
The portfolio is sorted
such that the probability of the performance of that portfolio is maximum at any time.
Experiments show that performance of a greedy strategy can be enhanced to near optimal over time.
%

%
The last two paragraphs cover recent literature in brief which explain previous strategies of combining different base methods.
The literature shows that they have performed well over different benchmarks. 
Our current settings do not capture any such ideas for combining different components of heuristics.  
A direct comparison with any of the above mentioned works is therefore out of scope for our current work.
This is because, we are more concerned about working with unit cost based POCL heuristics in isolation.
%
On the other hand, we suspect that many of these strategies, in some adapted form, 
would likely be beneficial in the POCL framework.
  
%
%
%
%
%
%
%
%
%
%
%
%
\section{Summary and Future Work}
We demonstrate the use of different regression models to combine different heuristic values to arrive at consistently better 
estimates over a range of planning domains and problems.
We extend some recent attempts to learn combinations of heuristic functions in 
state-space based planning to POCL planning.
We also show that the learned models can be further enhanced by an online error correction approach. 
In future we intend to explore online learning further, and continue our experiments with combining heuristic functions. 
We also aim to explore the use of an optimizing planner in tandem with bootstrapping methods. 
Apart from these, we will be giving a complete generalization of our current learning approaches for temporal planning and 
planning with deadlines.
%
%
%
%
%
\bibliography{formatting-instructions-latex}

\begin{thebibliography}{}

\bibitem[\protect\citeauthoryear{Arfaee, Zilles, and
  Holte}{2011}]{arfaee2011learning}
Arfaee, S.~J.; Zilles, S.; and Holte, R.~C.
\newblock 2011.
\newblock Learning heuristic functions for large state spaces.
\newblock {\em Artificial Intelligence} 175(16-17):2075--2098.

\bibitem[\protect\citeauthoryear{Benton, Coles, and Coles}{2012}]{BentonCC12}
Benton, J.; Coles, A.~J.; and Coles, A.
\newblock 2012.
\newblock Temporal planning with preferences and time-dependent continuous
  costs.
\newblock In {\em Proc. {ICAPS} 2012},  2--10.

\bibitem[\protect\citeauthoryear{Bishop}{2006}]{bishop2006pattern}
Bishop, C.~M.
\newblock 2006.
\newblock {\em Pattern {R}ecognition and {M}achine {L}earning}, volume~1.
\newblock Springer, Aug. 2006.

\bibitem[\protect\citeauthoryear{Coles \bgroup et al\mbox.\egroup
  }{2010}]{ColesCFL10}
Coles, A.~J.; Coles, A.; Fox, M.; and Long, D.
\newblock 2010.
\newblock Forward-chaining partial-order planning.
\newblock In {\em Proc. {ICAPS}}.

\bibitem[\protect\citeauthoryear{Domshlak, Karpas, and
  Markovitch}{2010}]{DomshlakKM10}
Domshlak, C.; Karpas, E.; and Markovitch, S.
\newblock 2010.
\newblock To max or not to max: Online learning for speeding up optimal
  planning.
\newblock In {\em Proc. {AAAI} 2010},  1071--1076.

\bibitem[\protect\citeauthoryear{Fawcett \bgroup et al\mbox.\egroup
  }{2014}]{FawcettVH0HL14}
Fawcett, C.; Vallati, M.; Hutter, F.; Hoffmann, J.; Hoos, H.~H.; and
  Leyton{-}Brown, K.
\newblock 2014.
\newblock Improved features for runtime prediction of domain-independent
  planners.
\newblock In {\em Proc. {ICAPS} 2014},  355--359.

\bibitem[\protect\citeauthoryear{Gerevini, Saetti, and
  Vallati}{2014}]{gerevini2014planning}
Gerevini, A.; Saetti, A.; and Vallati, M.
\newblock 2014.
\newblock Planning through automatic portfolio configuration: The {P}b{P}
  approach.
\newblock {\em JAIR 2014.}  639--696.

\bibitem[\protect\citeauthoryear{Hall \bgroup et al\mbox.\egroup
  }{2009}]{Hall2009}
Hall, M.; Frank, E.; Holmes, G.; Pfahringer, B.; Reutemann, P.; and Witten,
  I.~H.
\newblock 2009.
\newblock The {WEKA} data mining software: An update.
\newblock In {\em SIGKDD Explorations 2009}, volume~11.

\bibitem[\protect\citeauthoryear{Hall}{2000}]{hall1999correlation}
Hall, M.~A.
\newblock 2000.
\newblock Correlation-based feature selection for discrete and numeric class
  machine learning.
\newblock In {\em Proc. {ICML}}.

\bibitem[\protect\citeauthoryear{Haslum and
  Geffner}{2000}]{haslum2000admissible}
Haslum, P., and Geffner, H.
\newblock 2000.
\newblock Admissible heuristics for optimal planning.
\newblock In {\em Proc. {AIPS} 2000},  140--149.

\bibitem[\protect\citeauthoryear{Haslum}{2012}]{Haslum12}
Haslum, P.
\newblock 2012.
\newblock Incremental lower bounds for additive cost planning problems.
\newblock In {\em Proc. {ICAPS} 2012},  74--82.

\bibitem[\protect\citeauthoryear{Helmert and Domshlak}{2011}]{helmert2011lm}
Helmert, M., and Domshlak, C.
\newblock 2011.
\newblock {LM}-{C}ut: Optimal planning with the landmark-cut heuristic.
\newblock In {\em Seventh {IPC}}.

\bibitem[\protect\citeauthoryear{Helmert and
  Geffner}{2008}]{helmert2008unifying}
Helmert, M., and Geffner, H.
\newblock 2008.
\newblock Unifying the causal graph and additive heuristics.
\newblock In {\em Proc. ICAPS 2008}.

\bibitem[\protect\citeauthoryear{Helmert, R{\"o}ger, and
  Karpas}{2011}]{helmert2011fast}
Helmert, M.; R{\"o}ger, G.; and Karpas, E.
\newblock 2011.
\newblock Fast downward stone soup: A baseline for building planner portfolios.
\newblock In {\em Proc. {ICAPS PAL} 2011},  28--35.
\newblock Citeseer.

\bibitem[\protect\citeauthoryear{Helmert}{2006}]{helmert2006fast}
Helmert, M.
\newblock 2006.
\newblock The fast downward planning system.
\newblock {\em J. Artif. Intell. Res. {(JAIR)} 2006,} 26:191--246.

\bibitem[\protect\citeauthoryear{Hoffmann and Nebel}{2001}]{hoffmann2001ff}
Hoffmann, J., and Nebel, B.
\newblock 2001.
\newblock The {FF} planning system: Fast plan generation through heuristic
  search.
\newblock {\em {JAIR} 2001}.

\bibitem[\protect\citeauthoryear{Kvarnstr{\"{o}}m}{2011}]{Kvarnstrom11}
Kvarnstr{\"{o}}m, J.
\newblock 2011.
\newblock Planning for loosely coupled agents using partial order
  forward-chaining.
\newblock In {\em Proc. {ICAPS} 2011}.

\bibitem[\protect\citeauthoryear{Nguyen and
  Kambhampati}{2001}]{nguyen2001reviving}
Nguyen, X., and Kambhampati, S.
\newblock 2001.
\newblock Reviving partial order planning.
\newblock In {\em Proc. {IJCAI} 2001},  459--466.

\bibitem[\protect\citeauthoryear{N{\'u}{\~n}ez, Borrajo, and
  L{\'o}pez}{2015}]{NunezBL15}
N{\'u}{\~n}ez, S.; Borrajo, D.; and L{\'o}pez, C.~L.
\newblock 2015.
\newblock Sorting sequential portfolios in automated planning.
\newblock In {\em Proc. {IJCAI} 2015},  1638--1644.
\newblock AAAI Press.

\bibitem[\protect\citeauthoryear{Quinlan and
  others}{1992}]{quinlan1992learning}
Quinlan, J.~R., et~al.
\newblock 1992.
\newblock Learning with continuous classes.
\newblock In {\em Proc. Australian Joint Conference on Artificial
  Intelligence}, volume~92,  343--348.

\bibitem[\protect\citeauthoryear{Richter and Westphal}{2010}]{richter2010lama}
Richter, S., and Westphal, M.
\newblock 2010.
\newblock The {LAMA} planner: Guiding cost-based anytime planning with
  landmarks.
\newblock {\em {JAIR} 2010,} 39:127--177.

\bibitem[\protect\citeauthoryear{Richter, Helmert, and
  Westphal}{2008}]{richter2008landmarks}
Richter, S.; Helmert, M.; and Westphal, M.
\newblock 2008.
\newblock Landmarks revisited.
\newblock In {\em Proc. AAAI 2008}, volume~8,  975--982.

\bibitem[\protect\citeauthoryear{Richter, Westphal, and
  Helmert}{2011}]{richter2011lama}
Richter, S.; Westphal, M.; and Helmert, M.
\newblock 2011.
\newblock {LAMA} 2008 and 2011.
\newblock In {\em Seventh {IPC} 2011},  117--124.

\bibitem[\protect\citeauthoryear{Roberts \bgroup et al\mbox.\egroup
  }{2008}]{RobertsHWd08}
Roberts, M.; Howe, A.~E.; Wilson, B.; and desJardins, M.
\newblock 2008.
\newblock What makes planners predictable?.
\newblock In {\em Proc. ICAPS 2008},  288--295.
\newblock AAAI.

\bibitem[\protect\citeauthoryear{R{\"{o}}ger and Helmert}{2010}]{roger2010more}
R{\"{o}}ger, G., and Helmert, M.
\newblock 2010.
\newblock The more, the merrier: Combining heuristic estimators for satisficing
  planning.
\newblock In {\em Proc. {ICAPS} 2010},  246--249.

\bibitem[\protect\citeauthoryear{R{\"o}ger, Pommerening, and
  Seipp}{2014}]{rogerfast14}
R{\"o}ger, G.; Pommerening, F.; and Seipp, J.
\newblock 2014.
\newblock Fast downward stone soup 2014.
\newblock In {\em Eighth {IPC} 2014}.

\bibitem[\protect\citeauthoryear{Rousseeuw and
  Leroy}{2005}]{rousseeuw2005robust}
Rousseeuw, P.~J., and Leroy, A.~M.
\newblock 2005.
\newblock {\em Robust regression and outlier detection}, volume 589.
\newblock John Wiley \& Sons.

\bibitem[\protect\citeauthoryear{Samadi, Felner, and
  Schaeffer}{2008}]{samadi2008learning}
Samadi, M.; Felner, A.; and Schaeffer, J.
\newblock 2008.
\newblock Learning from multiple heuristics.
\newblock In {\em Proc. {AAAI} 2008},  357--362.

\bibitem[\protect\citeauthoryear{Sapena, Onaind{\i}a, and
  Torreno}{2014}]{sapena2014combining}
Sapena, O.; Onaind{\i}a, E.; and Torreno, A.
\newblock 2014.
\newblock Combining heuristics to accelerate forward partial-order planning.
\newblock In {\em Proc {ICAPS COPLAS} 2014}.

\bibitem[\protect\citeauthoryear{Seipp \bgroup et al\mbox.\egroup
  }{2015}]{SeippSHH15}
Seipp, J.; Sievers, S.; Helmert, M.; and Hutter, F.
\newblock 2015.
\newblock Automatic configuration of sequential planning portfolios.
\newblock In {\em Proc. {AAAI} 2015},  3364--3370.

\bibitem[\protect\citeauthoryear{Shekhar and Khemani}{2015}]{ShekharK15}
Shekhar, S., and Khemani, D.
\newblock 2015.
\newblock Improving heuristics on-the-fly for effective search in plan space.
\newblock In {\em {KI} 2015: Advances in Artificial Intelligence - 38th Annual
  German Conference on AI},  302--308.

\bibitem[\protect\citeauthoryear{Streeter and Smith}{2008}]{StreeterS08}
Streeter, M.~J., and Smith, S.~F.
\newblock 2008.
\newblock New techniques for algorithm portfolio design.
\newblock In {\em Proc. {UAI} 2008},  519--527.

\bibitem[\protect\citeauthoryear{Streeter, Golovin, and
  Smith}{2007}]{StreeterGS07}
Streeter, M.~J.; Golovin, D.; and Smith, S.~F.
\newblock 2007.
\newblock Combining multiple heuristics online.
\newblock In {\em Proc. {AAAI} 2007}.

\bibitem[\protect\citeauthoryear{Sutton}{1988}]{sutton1988learning}
Sutton, R.~S.
\newblock 1988.
\newblock Learning to predict by the methods of temporal differences.
\newblock {\em Machine Learning,} 3:9--44.

\bibitem[\protect\citeauthoryear{Thayer, Dionne, and
  Ruml}{2011}]{thayer2011learning}
Thayer, J.~T.; Dionne, A.~J.; and Ruml, W.
\newblock 2011.
\newblock Learning inadmissible heuristics during search.
\newblock In {\em Proc. {ICAPS} 2011}.

\bibitem[\protect\citeauthoryear{Virseda, Borrajo, and
  Alc{\'a}zar}{2013}]{us2013learning}
Virseda, J.; Borrajo, D.; and Alc{\'a}zar, V.
\newblock 2013.
\newblock Learning heuristic functions for cost-based planning.
\newblock In {\em Proc. {ICAPS PAL} 2013}.
\newblock Citeseer.

\bibitem[\protect\citeauthoryear{Younes and Simmons}{2003}]{younes2003vhpop}
Younes, H. L.~S., and Simmons, R.~G.
\newblock 2003.
\newblock {VHPOP:} {V}ersatile {H}euristic {P}artial {O}rder {P}lanner.
\newblock {\em {JAIR} 2003,} 20:405--430.

\end{thebibliography}
\bibliographystyle{aaai}
\section*{Appendix}
\subsection*{Proof of Theorem~1}
\subsubsection{Theorem 1.} 
\emph
{
For a given learned predictive model ($h$) and partial plan ($\pi_i$) in Figure~1 which leads to the solution plan ($\pi_{sol}$)
after certain refinement steps, the enhanced version of the predictive model ($h^e$) is,
\begin{equation}\label{eq2a:test}
		h^e(\pi_i) \ = \ h(\pi_{i}) \ + 
				\sum_{	\substack{\pi{'}~\text{{from}} \\ \pi_i \rightsquigarrow \pi_{sol}} } \epsilon_{h(\pi')}
\end{equation}
where $\pi_i$ $\rightsquigarrow$ $\pi_{sol}$ is a path in Figure~1 which captures each partial plan ($\pi'$) along 
the path between $\pi_i$ and $\pi_{sol}$.
The path includes $\pi_i$  and excludes the $\pi_{sol}$. The term $\epsilon_h$ is single-step-error associated with 
$\mathit{h}$ during refinement.
}
\begin{Proof} 
We use the principle of mathematical induction to prove this theorem.

\noindent $\mathit{\bold{Basis:}}$ We assume that $\pi_i$ needs only one refinement to become $\pi_{sol}$ that
would also be $\pi_i$'s best child.
Here, the best child always keeps the lowest estimate of requirement of new actions for its refinement among its siblings.
One possible refinement of $\pi_i$ is, $\pi_i \xrightarrow{R_i} \pi_{sol}$. 
Using Eq.~\eqref{eq2a:test}, we say,
\begin{eqnarray}\label{aeq2:test}	
		h^e(\pi_i) \ &=& \ h(\pi_{i}) \ +  \ \epsilon_{h(\pi_i)}
\end{eqnarray}	
The term $\epsilon_{h(\pi_i)}$ is the single-step-error associated with \emph{h} that estimates the total effort required for
$\pi_i$ to refine it completely. For unit cost refinements (cost($R_i$) = 1), $\epsilon_{h(\pi_i)}$ is computed as,
\begin{eqnarray}\label{aeq3:test}
	\epsilon_{h(\pi_i)} = (cost(R_i) + h(\pi_{i+1})) - h(\pi_{i})
\end{eqnarray}	
Here, the partial plan $\pi_{i+1}$ is also the $\pi_{sol}$, therefore h($\pi_{i+1}$) = 0. By using Eq.~\eqref{aeq2:test}
and Eq.~\eqref{aeq3:test} together, we get $h^e(\pi_i)$ = 1. Therefore, the base step holds.
%
%
%

%
In the base case, we assume that after refinement step $R_i$, to be the best child, there is no unsupported causal link present in 
$\pi_{i+1}$ and a threat (if any) will be resolved by the planner immediately to make it a solution plan. 
If there is an unsupported causal link then there must be an existing action in $\pi_{i+1}$ to support it. 
In this case, the estimate of requirement of new actions is still be 0.

\noindent $\mathit{\bold{Hypothesis:}}$ We select an arbitrary partial plan $\pi_{i+1}$ and assume that Eq.~\eqref{eq2a:test} holds for it.

\noindent $\mathit{\bold{Proof \ Step:}}$
Here, we show that Eq.~\eqref{eq2a:test} holds for $\pi_i$ too.
\begin{eqnarray*}
		h^e(\pi_i) &=& cost(R_i)  +  h^e(\pi_{i+1}) \\						
						&=&  cost(R_i)  +  h(\pi_{i+1}) +  
										\sum_{\substack{\pi'~{\textit{from}} \
										\pi_{i+1} \rightsquigarrow \pi_{sol}}} \epsilon_{h(\pi')} \\
											&& \text{By the induction hypothesis.}
										\\ &=& h(\pi_{i})  +  \epsilon_{h(\pi_i)}
							  			 +  \sum_{\substack{\pi'~\textit{{from}} 
							  			 \ \pi_{i+1} \rightsquigarrow \pi_{sol}}} \epsilon_{h(\pi')} 
						  								~~\text{by Eq.~}\eqref{aeq3:test}.
							\\ &=& 	h(\pi_{i})  + \sum_{\substack{\pi'~\textit{{from}} \ \pi_{i} \rightsquigarrow 
							\pi_{sol}}} \epsilon_{h(\pi')}
\end{eqnarray*}
Therefore, the the relationship holds for the parent partial plan $\pi_i$ as well.
Thus, by induction for all partial plans $\pi$, our assumption is correct. $\hspace*{143pt}$ \tiny $\blacksquare$ 
\end{Proof}
\end{document}